\pdfoutput=1
\documentclass[runningheads]{llncs}
\usepackage{graphicx}
\usepackage{hyperref}

\usepackage{amsmath}
\usepackage{amssymb}
\usepackage{caption}
\usepackage{float}
\usepackage[misc]{ifsym}

\begin{document}
%

\title{IDPL: Intra-subdomain adaptation adversarial learning segmentation method based on Dynamic Pseudo Labels}
\titlerunning{IDPL}
%

\author{Xuewei Li\inst{1,2,3}\orcidID{0000-0002-5330-7298} \and
Weilun Zhang\inst{4} \and
Jie Gao\inst{1,2,3} \and
Xuzhou Fu\inst{1,2,3} \and
Jian Yu\inst{1,2,3}$^{(\textrm{\Letter})}$
}
\authorrunning{X. Li et al.}
%
\institute{College of Intelligence and Computing, Tianjin University, Tianjin, China. \and
Tianjin Key Laboratory of Cognitive Computing and Application, Tianjin, China. \and
Tianjin Key Laboratory of Advanced Networking, Tianjin, China.
\email{\{lixuewei,gaojie,fuxuzhou,yujian\}@tju.edu.cn}\and
Tianjin International Engineering Institute, Tianjin University, Tianjin, China. 
\email{zhangweilun@tju.edu.cn}\\
$^{\textrm{\Letter}}$ Corresponding Author
}
\toctitle{IDPL: Intra-subdomain adaptation adversarial learning segmentation method based on Dynamic Pseudo Labels}
\tocauthor{Xuewei ~ Li\inst{1,2,3}\orcidID{0000-0002-5330-7298},
Weilun ~ Zhang\inst{4},
Jie ~ Gao\inst{1,2,3},
Xuzhou ~ Fu\inst{1,2,3},
Jian ~ Yu\inst{1,2,3}$^{(\textrm{\Letter})}$
}
\maketitle              

\begin{abstract}
Unsupervised domain adaptation(UDA) has been applied to image semantic segmentation to solve the problem of domain offset. However, in some difficult categories with poor recognition accuracy, the segmentation effects are still not ideal. To this end, in this paper, Intra-subdomain adaptation adversarial learning segmentation method based on Dynamic Pseudo Labels(IDPL) is proposed. The whole process consists of 3 steps: Firstly, the instance-level pseudo label dynamic generation module is proposed, which fuses the class matching information in global classes and local instances, thus adaptively generating the optimal threshold for each class, obtaining high-quality pseudo labels. Secondly, the subdomain classifier module based on instance confidence is constructed, which can dynamically divide the target domain into easy and difficult subdomains according to the relative proportion of easy and difficult instances. Finally, the subdomain adversarial learning module based on self-attention is proposed. It uses multi-head self-attention to confront the easy and difficult subdomains at the class level with the help of generated high-quality pseudo labels, so as to focus on mining the features of difficult categories in the high-entropy region of target domain images, which promotes class-level conditional distribution alignment between the subdomains, improving the segmentation performance of difficult categories. For the difficult categories, the experimental results show that the performance of IDPL is significantly improved compared with other latest mainstream methods.

\end{abstract}

\section{Introduction}

The domain shift problem in UDA manifests itself as the inter-domain discrepancy problem between synthetic images and real images in semantic segmentation task, making difficult to generalize the model trained with synthetic data to real data. Previous UDA methods~\cite{Tsai_2018_CVPR,Vu_2019_CVPR} have done a lot of works to reduce the domain shift, but there are still some problems with existing methods: First of all, most of the self-training methods based on pseudo labels select a fixed threshold to filter all pseudo labels with high confidence~\cite{Zou_2018_ECCV,Zou_2019_ICCV}. The model pay more attention to the conditional distribution alignment of the samples in easy categories between the two domains, resulting in the samples in difficult categories with high entropy may be discarded; Moreover, most methods adopt the strategy of global threshold~\cite{zheng2019unsupervised,Lian_2019_ICCV}, ignoring the separate consideration of different classes, which increases the risk of class imbalance in the generated pseudo labels.These problems lead to low accuracy of generated pseudo labels.

\begin{figure}[ht]

\centering
\includegraphics[width=1\linewidth]{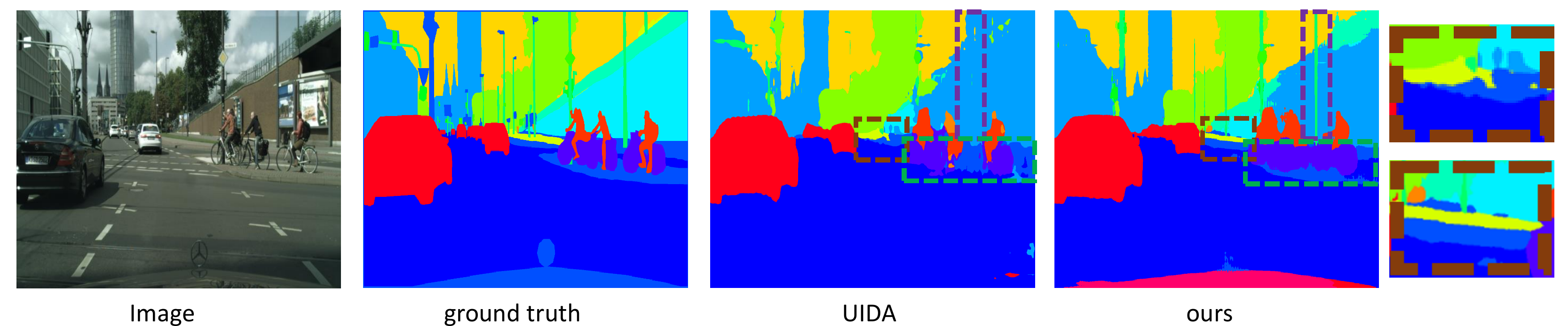}
\caption{\ Comparison of pseudo label results between IDPL and similar method.}
\label{fig1}
\end{figure}

In addition, to alleviate the distribution gap between data within the target domain, the unsupervised intra-domain adaptation methods are proposed. For example, UIDA~\cite{Pan_2020_CVPR} considers the global feature map of samples during adversarial training between the easy and difficult subdomains. However, not all spatial regions maintain high entropy for difficult samples. Using global features for adversarial learning, the feature extraction of difficult samples in low-entropy regions may be affected, causing negative transfer. Therefore it is not reliable to use the global entropy of the predicted probability maps to divide easy/difficult subdomains. The results are shown in {\bf Fig.~\ref{fig1}}, for "difficult" categories (such as "pole" in purple boxes, "bike" in green boxes, and "sidewalk" in brown boxes), this method does not work well.

Inspired by UIDA~\cite{Pan_2020_CVPR}, this paper proposes a more stable domain adaptation method to achieve intra-subdomain adversarial training, namely Intra-subdomain adaptation adversarial learning method based on Dynamic Pseudo Labels (IDPL). The method consists of 3 parts: Firstly, in order to improve the pseudo labels quality of intra-domain adversarial learning, the instance-level pseudo labels dynamic generation module is proposed. The threshold is dynamically adjusted for different semantic classes of each image, so that the model pays more attention to the high-entropy regions in the image; Then, the subdomain classifier module based on instance confidence is constructed to realize the hierarchical division from easy instances/difficult instances to easy subdomain/difficult subdomain; Finally, on the basis of the divided two subdomains, the subdomain adversarial learning module based on self-attention is constructed, and the multiple discriminator head structure is introduced to mine the class information contained in the high-entropy or entropy fluctuation regions, so as to more accurately narrow the intra-domain differences. As shown in {\bf Fig.~\ref{fig1}}, the pseudo labels of IDPL are significantly better than the comparison method, especially in these "difficult" categories.

\graphicspath{{figs/}}

The main contributions of this paper are summarized as follows:

(1)The instance-level pseudo labels dynamic generation module is proposed, which aims to dynamically adjust the global threshold of each class according to the instance prediction of each class, fuse local and global information to guide the model to generate high-quality pseudo labels.

(2)The intra-subdomain adversarial learning module based on instance confidence is proposed. The confidence coefficient is constructed according to the relative proportion of easy and difficult instances, thus the target domain can be accurately divided into image-level easy and difficult subdomains; Guide the model to generate subdomain invariant features through intra-domain adversarial training at the semantic class level, so that the model pays more attention to the more difficult categories, and the influence of concentrated high-entropy regions on segmentation results can be alleviated more effectively.

\section{Method}

\begin{figure*}[!htbp]

\centering
\includegraphics[width=1\linewidth]{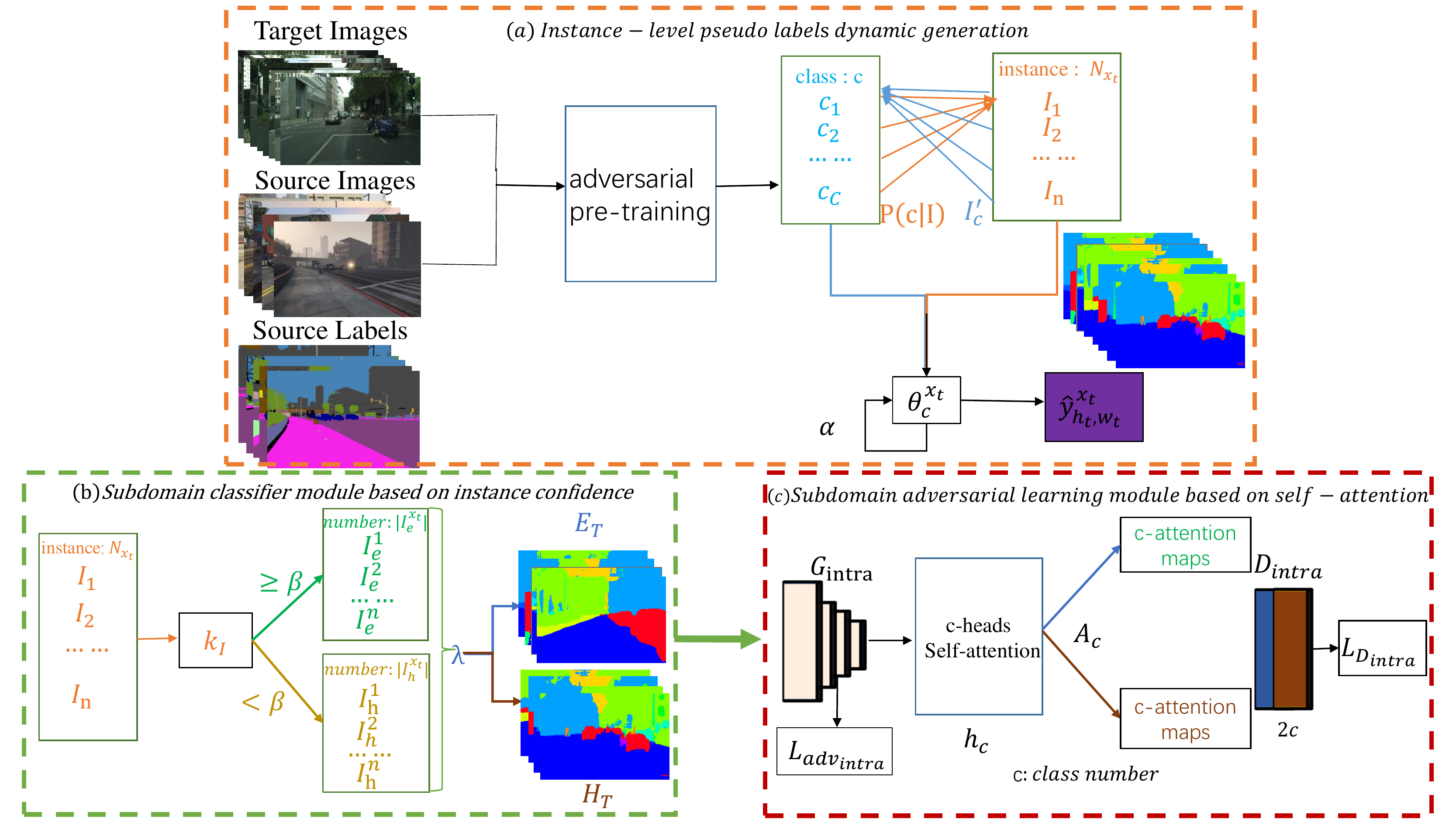}
\caption{\ The framework of the method proposed in this paper. }
\label{fig2}
\end{figure*}

The overall framework is shown in {\bf Fig.~\ref{fig2}}. The method can be divided into 3 steps. Firstly, momentum is introduced for each instance to gradually update the threshold for each class, generating high-quality pseudo labels. Secondly, the confidence of each instance is calculated, and they are classified, then the target domain is split according to the relative proportion of the two types of instances. Thirdly, the self-attention heads are used to apply self-attention to each class of the two subdomains separately, and the self-attention maps are used as the weight of each class to conduct inter-class subdomain adversarial training.

\subsection{Instance-level pseudo labels dynamic generation module(PLDG)}

First, the pre-trained segmentation model G is used for preliminary prediction of the target domain image, and the predicted probability maps $M_{T}=G\left(X_{T}\right)$ is obtained. Then, softmax is used to calculate the class probability prediction of each pixel position $\left(h_{t}, w_{t}\right)$ in the image $X_{T}$, as shown in Eq. (\ref{e1}):

\begin{equation}
p_{i, h_{t}, w_{t}}^{x_{t}}=\frac{e^{m_{t}^{\left(h_{t}, w_{t}, i\right)}}}{\sum_{j=1}^{C} e^{m_{t}^{\left(h_{t}, w_{t}, j\right)}}}, m_{t} \in M_{t}
\label{e1}
\end{equation}

\noindent Where $i$ represents a specific semantic class, $i \in\{0,1, \ldots, C\}$; The number with the largest value along the channel direction of a certain pixel position $\left(h_{t}, w_{t}\right)$ represents the class prediction probability of this pixel, and the corresponding channel subscript represents the predicted class. In the process of generating pseudo labels, the weight parameters of the generator network are fixed. The pseudo labels generation can be expressed by Eq. (\ref{e2}):

\begin{equation}
\min _{\hat{Y}_{T}}-\sum_{h_{t}, w_{t}} \sum_{c} \hat{y}_{h_{t}, w_{t}}^{x_{t}} \log \frac{p_{c, h_{t}, w_{t}}^{x_{t}}}{\theta_{c}^{x_{t}}}
\label{e2}
\end{equation}

\noindent Where $\theta_{c}^{x_{t}}$ represents the threshold of the $C$-th class at the pixel position $\left(h_{t}, w_{t}\right)$, and $\hat{y}_{h_{t}, w_{t}}^{x_{t}}$ indicates the pseudo labels of a vector of C classes. 

Considering the problem of unbalanced sample number corresponding to different semantic classes, a threshold $\theta_{c}$ about class is introduced, which is used to provide a specific threshold for each of c different semantic classes in a batch, which is denoted as the global class thresholds. In order to generate high-quality pseudo labels map $\hat{y}_{h_{t}, w_{t}}^{x_{t}}$, this module aims to generate an adaptive threshold $\theta_{c}^{x_{t}}$ for each class c. The update process of the dynamic threshold of pseudo labels is described in detail below.

First, for the pseudo labels activation region $H_{i}=\left\{(h, w) \mid \hat{y}_{h_{t}, w_{t}}^{x_{t}}=1\right\}, i \in\{0,1, \ldots, C\}$ in a specific channel, the feature vector $F_{i}=\left\{f_{x, y} \mid(x, y) \in H_{i}\right\}$ of the corresponding region in the original feature map is selected. At this point, with the help of the clustering algorithm, the features of this region are separated into different instances, and the total number of instances sampled from a certain image sample $x_{t}$ is denoted as $N_{x_{t}}$. Global average pooling is introduced to obtain the embedding vector of each different instance, denoted as $I_{n}$, where $n \in\left\{1,2, \ldots, N_{x_{t}}\right\}$. Then, instances of all classes are more appropriately matched based on their similarity between the global and the local, in which all instances need to be dynamically classified. In addition, the method introduces a linear layer and Softmax activation function as a classifier for each instance feature $I$. The class prediction probability $p(\mathrm{c} \mid \mathrm{I})$ of each instance is expressed by Eq. (\ref{e3}):

\begin{equation}
p(c \mid I)=\operatorname{softmax}\left(W_{c}^{T} I\right)
\label{e3}
\end{equation}

\noindent Where $W_{c}$ is the trainable parameter of the class classifier. Then, a class sampling weight $I_{c}^{\prime}$ is set, and the sampling process is shown in Eq. (\ref{e4}):

\begin{equation}
I_{c}^{\prime}=\frac{1}{h_{t} \times w_{t}} \sum_{h_{t}, w_{t}} p_{c, h_{t}, w_{t}}^{x_{t}} \times L_{c}, \forall x_{t} \in X_{T}
\label{e4}
\end{equation}

\noindent Where $L_{c}$ represents a set of several instance features, expressed as \\$L_{c}=\left\{I_{n} \mid n \in \\ \left\{1,2, \ldots, N_{x_{t}}\right\}\right\}$, $\times$ represents the sampling operation for matching and fusion.

Finally, when pseudo labels are generated, a global information is retained after iteration for each instance $I$; A local class threshold $\hat{\theta}_{c}^{x_{t}}$ is set for each image sample in the current batch, which is obtained by averaging the predicted classification of all instances in each class. The semantic class threshold at the image level can be obtained by fusing the local class threshold $\hat{\theta}_{c}^{x_{t}}$ and the global class thresholds $\theta_{c}$ initialized by the current batch in a certain proportion, thereby gradually generating high-quality pseudo labels. The process of threshold fusion is shown in Eq. (\ref{e5}):

\begin{equation}
\theta_{c}^{x_{t}}=\alpha \theta_{c}+(1-\alpha) \hat{\theta}_{c}^{x_{t}}
\label{e5}
\end{equation}

$\alpha$ is the momentum factor used to hold the global class thresholds, with a range of (0,1). Within a certain range, as $\alpha$ gets larger, the dynamic update of $\theta_{c}^{x_{t}}$ becomes smoother and smoother. In this way, the global class thresholds and the class thresholds corresponding to the local instance are matched and fused to realize the adaptive matching between the global and local classes, so the dynamic iterative updating of the image-level class threshold is realized.

\subsection{The subdomain classifier module based on instance confidence(SCIC)}

In order to capture the high-entropy regions information and the difficult samples are accurately mined, this module can achieve more reliable image-level subdomain division of the target domain, thereby providing better guidance for reducing intra-domain differences. 

Specifically, $\beta$ is selected as the division threshold of the instances. Combined with the dynamic threshold $\theta_{c}^{x_{t}}$ of each class and the prediction probability $p(\mathrm{c} \mid \mathrm{I})$ of each instance, the confidence of each instance can be calculated, as shown in Eq. (\ref{e6}):

\begin{equation}
k_{I}=p(c \mid I) \theta_{c}^{x_{t}}
\label{e6}
\end{equation}

Defining the set of easy instances as $I_{e}$ and the set of difficult instances as $I_{h}$. When $k_{I} \geq \beta$, the instance is divided into $I_{e}$, otherwise into $I_{h}$.

$\lambda$ is selected as the split ratio threshold of the target domain. Defining the easy subdomain as $E_{T}$ and the difficult subdomain as $H_{T}$. Let $\left|I_{e}^{x_{t}}\right|$ and $\left|I_{h}^{x_{t}}\right|$ represent the number of instances of $I_{e}$ and $I_{h}$ in image $x_{t}$ respectively. When $\left|I_{e}^{x_{t}}\right| /\left(\left|I_{h}^{x_{t}}\right|+\left|I_{e}^{x_{t}}\right|) \geq \lambda\right.$, the image is divided into $E_{T}$; otherwise into $H_{T}$.

\subsection{The subdomain adversarial learning module based on self-attention(SASA)}

In order to reduce sample differences within the target domain, adversarial learning is introduced in this section to confuse the sample spatial distribution of the easy subdomain and the difficult subdomain. In addition, in order to further mine the correlation between pixel-level same semantic regions in target domain images, the self-attention mechanism is introduced for the adversarial learning of easy/difficult subdomains to train by class. Since the two subdomains share the same c semantic classes, the multi-head self-attention module introduces the self-attention mechanism for each class separately, achieve promoting the alignment of conditional distribution at the semantic class level. 

The predicted probability maps obtained by the easy/difficult subdomains through the segmentation network are input into the self-attention module with c heads $h_{c}$, and the multi-head self-attention module generates c class self-attention maps $A_{1}, A_{2}, \cdots, A_{C}$ (each class corresponds to a self-attention map).

To focus on high-entropy regions, this paper improves the binary classification head of the traditional discriminator, and allocates a discriminator for each semantic class, namely c parallel binary discriminators, which forms a new class-level intra-domain discriminator $D_{intra}$.

The training purpose of $D_{intra}$ is to distinguish whether the image belongs to the easy subdomain or the difficult subdomain, as shown in Eq. (\ref{e7}):

\begin{equation}
\begin{gathered}
\min _{D_{intra}} L_{D_{\text {intra }}}\left(E_{T}, H_{T}\right)= \\
-\sum_{c} Q_{e, c}(1-d) \log p\left(d=0, c \mid f_{e}\right)-\sum_{c} Q_{h, c} d \log p\left(d=1, c \mid f_{h}\right)
\end{gathered}
\label{e7}
\end{equation}

\noindent where $Q_{e, c}$ and $Q_{h, c}$ are the weights of class c of images in the “easy” subdomain and the “difficult” subdomain respectively, $Q \in R^{1 \times C}$. $f_{e}$ and $f_{h}$ represent the features extracted by $F_{intra}$ from the images of two subdomains respectively, $d$ represents the domain variable, where 0 represents the easy subdomain and 1 represents the difficult subdomain. $p(d \mid f)$ represents the probability of the discriminator $D_{intra}$ output. Similarly, to obtain subdomain-invariant features, the discriminator $D_{intra}$ is confused so that it cannot distinguish between easy and difficult subdomains. As shown in Eq. (\ref{e8}):

\begin{equation}
\label{e8}
\min _{G_{intra}} L_{a d v_{\text {intra }}}\left(H_{T}\right)=-\sum_{c} Q_{h, c} \log p\left(d=0, c \mid f_{T}\right)
\end{equation}

\noindent $L_{a d v_{\text {intra }}}\left(H_{T}\right)$ aims to maximize the probability that the "difficult" subdomain feature is regarded as the "easy" subdomain feature without damaging the relationship between features and classes.

\section{Experiment}

\subsection{Experimental setup}

{\bf Dataset:}IDPL is evaluated on commonly used semantic segmentation tasks from synthetic domain to real domain: GTA5~\cite{richter2016playing} to Cityscapes~\cite{cordts2016cityscapes} and SYNTHIA~\cite{ros2016synthia} to Cityscapes. The GTA5 dataset has 24,966 images from GTA5, and 19 classes with urban scenes. The SYNTHIA dataset includes 9400 images and 16 classes with urban scenes. The Cityscapes dataset is divided into training set, validation set and test set. Following the standard protocol in~\cite{Tsai_2018_CVPR}, using the training set of 2975 images as the target domain dataset and using the validation dataset to evaluate the model using IoU and mIoU.

\noindent {\bf Implementation details:}This paper uses PyTorch for training and inference on a single NVIDIA RTX TiTan/24GB. For fair comparison, DeepLabv2~\cite{chen2017deeplab} is used as the basic segmentation network, ResNet-101~\cite{he2016deep} and VGG16~\cite{simonyan2014very} as the backbone network. The training images are cropped randomly and resized to $1024 \times 512$. The initial learning rate is set to $2 \times 10^{-4}$ and reduced according to the “poly” learning rate strategy with a power of 0.9. To train the discriminator, the Adam optimizer is used with $\beta_{1}=0.9$, $\beta_{2}=0.99$ and initial learning rate of $10^{-4}$. Through experiments, the optimal hyperparameters obtained are set as $\alpha=0.9$, $\beta=0.6$, $\lambda=0.7$.

\subsection{Ablation experiment}

In order to understand the hyperparameter values and the impact of each module, we carry out ablation experiments using the ResNet-101 backbone in the GTA5 $\rightarrow$ Cityscapes task.  

\subsubsection{Module ablation experiment}

{\bf Table~\ref{tab1}} shows the results that verify the effectiveness of different modules in the method. Define PT as the model obtained by inter-domain adversarial pre-training using AdvEnt~\cite{Vu_2019_CVPR}. PT is the subsequent baseline model, mIoU=41.0\%. PT+PLDG is 5.7\% higher than PT. The method generates an adaptive semantic class threshold for each class by fusing the global and local information of each class instance by instance, achieving the clustering of the same class samples of inter-domain on the basis of edge alignment, which is feasible to improve the quality of pseudo labels. PT+PLDG+SASA, self-attention adversarial training is carried out by class in the two subdomains respectively, 7.5\% higher than PT+PLDG. Subdomain adversarial training based on high-quality pseudo labels can focus on mining difficult categories in high-entropy regions, accurately guide the model to perform intra-domain adaptation in high-entropy and entropy fluctuation regions, and further improve the performance of the model.

\begin{table}[]
\caption{\ Ablation experiments where each module is added individually.}
\centering
\begin{tabular}{ccccc}
\hline
PT & PLDG & SASA & mIoU & $\Delta$ \\ \hline
\checkmark   &                &                 & 41.0          &  -     \\
\checkmark   &\checkmark      &                 & 46.7          &  +5.7     \\
\checkmark   & \checkmark     &\checkmark       & {\bf54.2}     &  +7.5     \\ \hline

\end{tabular}
\label{tab1}
\end{table}

{\bf Table~\ref{tab2}} shows the results of ablation experiments by replacing each module in the framework with a module with similar function in other methods.

{\bf PLDG module:} This part verifies the pseudo labels generated by AdvEnt~\cite{Vu_2019_CVPR} proposed in UIDA~\cite{Pan_2020_CVPR}, when these pseudo labels are used in the subsequent tasks of the proposed method, which is 4.4\% lower than the final result of IDPL. This indicates that PLDG module sets a threshold for different classes respectively, and dynamically fine-tunes each semantic class threshold according to the confidence prediction of each instance with the cyclic training of the model, generating higher quality pseudo labels. 

{\bf SCIC module:} This part randomly selects half of the images to be split into the "easy" subdomain and the rest into the "difficult" subdomain. The two subdomains are input into the subsequent module for experiments, which is 2.8\% lower than the final result of ours. Therefore, this result indicates that selecting a fixed value as the threshold for domain separation cannot well perform targeted filtering for useful labels of each class. While SCIC module dynamically split the target domain into easy/difficult subdomains by using two hyperparameters for different datasets and tasks by considering the relative proportion of easy and difficult instances reasonably, implements hierarchical division of easy/difficult instances to easy/difficult subdomains, thus improve the model performance under different tasks. In particular, on the premise of not affecting the performance of "easy" categories, the goal of improving the segmentation accuracy of "difficult" categories is achieved by increasing the number of "difficult" categories as much as possible.

\begin{figure*}[ht]

\centering
\includegraphics[width=1\linewidth]{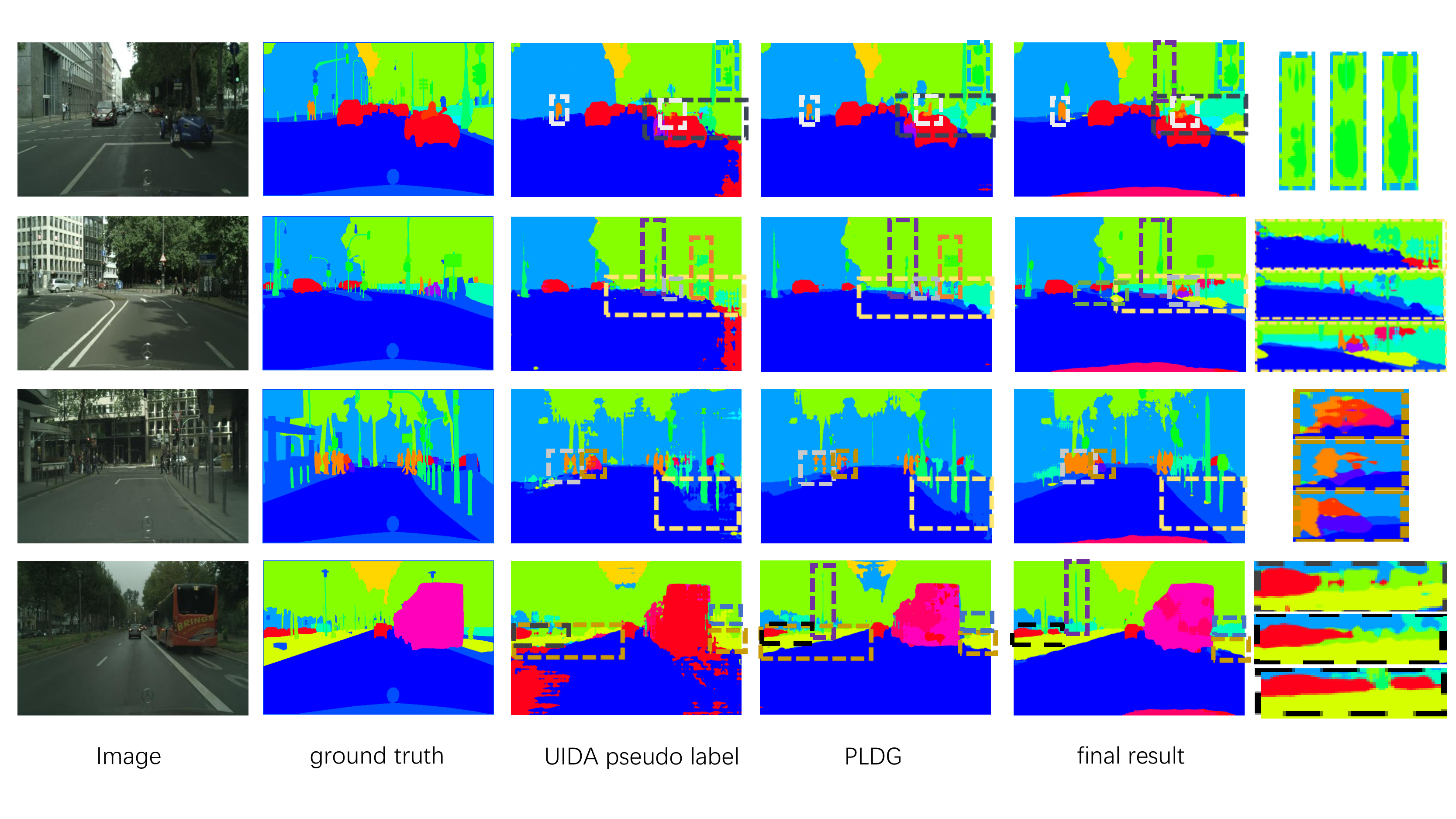}
\caption{\ Visualization results on the GTA5 $\rightarrow$ Cityscapes task.}
\label{fig3}
\end{figure*}

{\bf SASA module:} Compared with the improvement work of AdvEnt~\cite{Vu_2019_CVPR} in UIDA~\cite{Pan_2020_CVPR}, that is, using the common generator and discriminator for intra-domain adversarial learning, which is 1.9\% lower than the final result of ours. This part conducts intra-domain adversarial training for each class in the two subdomains with the help of the self-attention mechanism. The C self-attention maps generated by the multi-head self-attention mechanism assign different weights to different classes, which solves the problem that UIDA~\cite{Pan_2020_CVPR} ignores the class information contained in regions with high-entropy or entropy fluctuation. The final visualization results are shown in {\bf Fig.~\ref{fig3}}, compared with UIDA~\cite{Pan_2020_CVPR} and PLDG module, IDPL improves the quality of generated pseudo labels, especially for the “difficult” categories.(Such as “sign” and “light” in blue boxes in the first image; “sidewalk”, “fence” and "mbike" in yellow boxes in the second image; “rider” and “bike” in brown boxes in the third image; “fence” in blue boxes in the fourth image.)

\begin{table}[]
\caption{\ Ablation experiments of each module is replaced by a similar module.}
\centering
\scalebox{0.9}{
\begin{tabular}{lcccccc}
\hline
Method               & PT & PLDG & SCIC & SASA & mIoU & $\Delta$ \\ \hline
IDPL(ours)  & \checkmark   & \checkmark     & \checkmark    &\checkmark       & {\bf54.2}     & -      \\
General pseudo label & \checkmark   &      & \checkmark    &\checkmark       & 49.8     &  -4.4    \\
Random Select        &\checkmark    & \checkmark     &     &\checkmark       & 51.4     &  -2.8     \\
UIDA{\cite{Pan_2020_CVPR}}         &\checkmark    & \checkmark     & \checkmark    &       & 52.3     &  -1.9      \\ \hline
\end{tabular}}
\label{tab2}
\end{table}

\subsubsection{Model parameter analysis}

In this section, sensitivity analysis will be conducted on the hyperparameters $\alpha$, $\beta$, $\lambda$ involved in Chapter 2.

{\bf $\alpha$:} In PLDG module, when the global iteration information momentum $\alpha = 0$, it means that there is only prior information for each class of the instance, and there is no global consideration by class. When iterating for each instance, with the increase of $\alpha$, the weight of global information is gradually increasing, and the weight of local information is gradually decreasing. This causes the threshold for each class to update more and more slowly, as shown in {\bf Table~\ref{tab3}}. When $\alpha=0.9$, on the basis of retaining the vast majority of the global information, the local information is slowly integrated, so as to better combine the global class information and the local instance information, and achieve the optimal threshold selection of each class at the image level. It has good adaptability to some difficult categories with few samples, and obtains more diversified information, which makes the model obtain the best performance.

\begin{table}
\setlength{\abovecaptionskip}{0pt}
\setlength{\belowcaptionskip}{10pt}
\caption{\ The study of the hyperparameter $\alpha$ for preserving global information.}
\begin{center}
\begin{tabular}{c}
\hline
\qquad\qquad\;\!\!GTA5 $\longrightarrow$ Cityscapes\qquad\qquad\;\!\!\!\\
\end{tabular}

\begin{tabular}{cccccccc}
\hline
$\alpha$    & 0 & 0.5 & 0.6 & 0.7 & 0.8 & 0.9 & 1.0 \\
mIoU & 41.7  & 47.7   & 49.3    & 50.8    & 53.2    & {\bf54.2}    & 53.8    \\ \hline
\end{tabular}
\end{center}
\label{tab3}
\end{table}

$\beta$ and $\lambda$ are selected 10 values from $[0.1,1.0]$ with a step size of 0.1, and the experimental results of 100 different combinations are presented in a three-dimensional stereogram. The intersection point is the mIoU of the corresponding combination. The darker the red is, the larger the value is.

\begin{figure}[ht]
\centering
\includegraphics[width=1\linewidth]{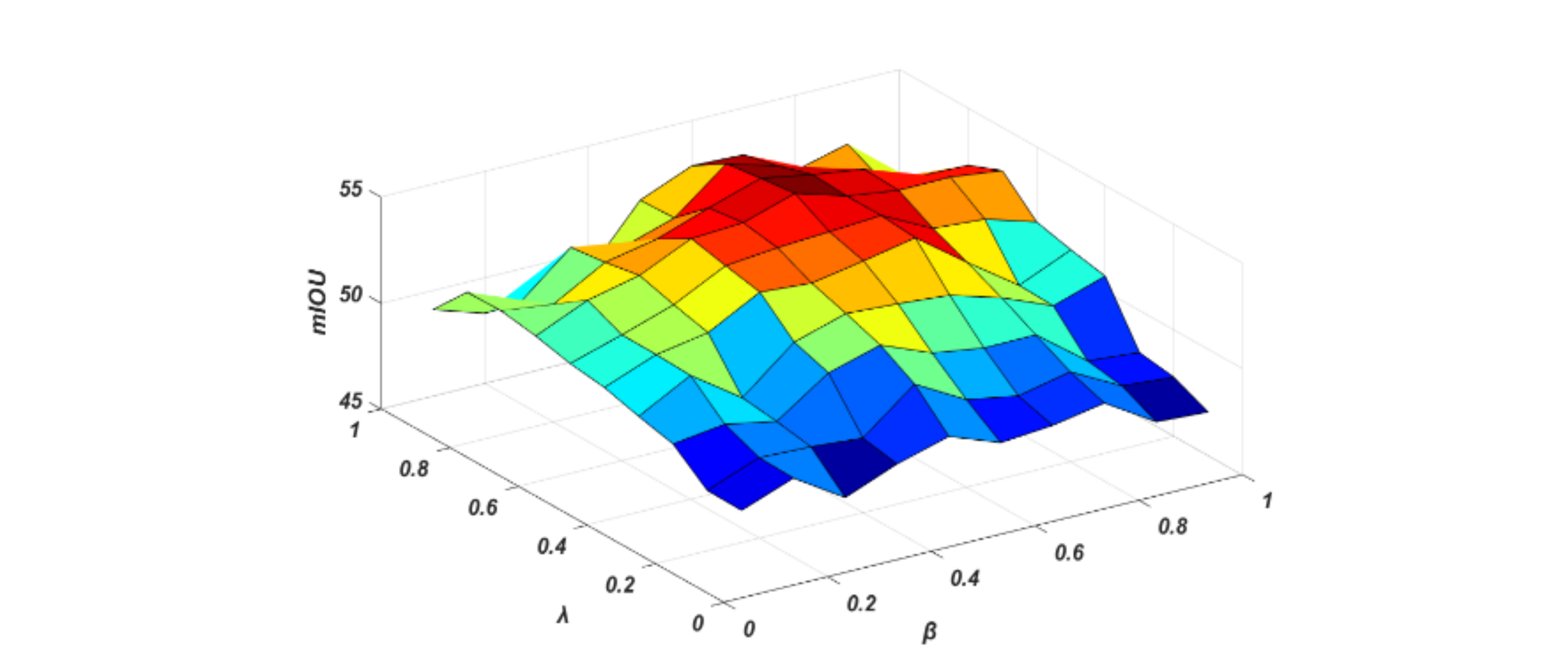}
\caption{\ The study of the hyperparameters $\beta$ and $\lambda$.}
\label{fig4}
\end{figure}

{\bf $\beta$:} When $\beta$ is small, most of the instances are divided into "easy" instances set, and there are not enough "difficult" instances to train the model; As shown in {\bf Fig.~\ref{fig4}}, as $\beta$ increases, the threshold for instances division into "easy" is increasing, the number of "difficult" instances is increasing, and the performance of the model is also increasing continuously; Until $\beta=0.6$, the optimal relative allocation ratio of the two types of instances can be obtained, and when $\lambda=0.7$, the best mIoU is reached; As $\beta$ continues to increase, the proportion of "difficult" instances keeps increasing, resulting in the model overfitting to "difficult" instances. At the same time, the number of "easy" instances is decreasing, and there are not enough "easy" instances to train the model, resulting in the decline of mIoU .

{\bf $\lambda$:} As $\lambda$ increases, the number of images in the "easy" subdomain is gradually decreasing, more and more images are divided into the "difficult" subdomain. Since most of the classes in the "difficult" subdomain belong to "difficult" categories, the model performance is gradually improved; The optimal value is $\lambda=0.7$, which realizes that the relatively difficult samples contained in the concentrated high-entropy region are divided into the "difficult" subdomain. At this point, the number of images in the "difficult" subdomain satisfies the training needs of the method for the "difficult" categories. As $\lambda$ further increases, the number of images in the "difficult" subdomain is further increasing. Since most of the classes in the "difficult" subdomain are "difficult" categories, the model lacks a sufficient number of "easy" categories to make mIoU drop.

\subsection{Comparative Experiment }

This section analyzes the quantitative results of IDPL and compares them with other mainstream methods. The results are shown in {\bf Table~\ref{tab4}} and {\bf Table~\ref{tab5}}. For fair comparison, all results are derived from single-scale inference. 

IDPL outperforms other mainstream methods in more than half of the classes on both backbones for two tasks, and mIoU also reaches a high level, and mIoU also reaches a high level. Especially in the "difficult" categories such as “fence” and “pole”, IDPL outperforms the others. On the one hand, these performance improvements show that the adaptive processing of classes is effective, and the dynamically generated pseudo labels can generate adaptive threshold for each class; On the other hand, when dividing the target domain, the method increases the number of images in the "difficult" subdomain as much as possible, and gives a higher weight to the "difficult" categories in the subdomain adversarial learning, adding more information of the "difficult" categories. Compared with the similar method UIDA~\cite{Pan_2020_CVPR} which adopts intra-domain adversarial learning, IDPL improved by 7.9\% and 9.2\% in mIoU in two tasks. In terms of specific classes, there is a significant improvement in almost all classes, especially in the difficult categories such as “fence” and “sign”, the results of IDPL have been doubled. This proves that compared with some methods that only consider global feature alignment, IDPL considers the differences of different classes, realizes feature alignment at the class level, and solves the problem of class mismatch after domain alignment. Compared with CBST~\cite{Zou_2018_ECCV} and DAST~\cite{yu2021dast}, which also use the discriminator, IDPL assigns higher weights to difficult categories through self-attention mechanism at the class level and multiple processing for difficult categories, thus achieving a great performance improvement.

\begin{table*}[]
\caption{\ IoU (\%) comparison for each class on GTA5 $\rightarrow$ Cityscapes adaptation, evaluated on Cityscapes validation set.}
\centering
\scalebox{0.6}{
\begin{tabular}{lcccccccccccccccccccc}
\hline
\multicolumn{1}{l|}{Method}      & road & sidewalk & building & wall & fence & pole & light & sign & veg  & terrain & sky  & person & rider & car  & truck & bus  & train & mbike & \multicolumn{1}{c|}{bike} & mIoU \\ \hline
VGG-16                  &      &       &       &      &       &      &       &      &      &      &      &      &      &      &       &      &       &      &                           &      \\ \hline
\multicolumn{1}{l|}{Source Only} & 60.7 & 13.7  & 56.9  & 12.9 & 20.1  & 19.0 & 15.4  & 6.5  & 77.7 & 16.2 & 56.8 & 40.0 & 3.3  & 63.6 & 15.3  & 9.5  & 0.0   & 8.1  & \multicolumn{1}{c|}{0.1}  & 26.1 \\
\multicolumn{1}{l|}{CBST\cite{Zou_2018_ECCV}}        & 90.4 & {\bf50.8}  & 72.0  & 18.3 & 9.5   & 27.2 & 28.6  & 14.1 & 82.4 & 25.1 & 70.8 & 42.6 & 14.5 & 76.9 & 5.9   & 12.5 & 1.2   & 14.0 & \multicolumn{1}{c|}{\bf28.6} & 36.1 \\
\multicolumn{1}{l|}{AdvEnt\cite{Vu_2019_CVPR}}      & 86.9 & 28.7  & 78.7  & 28.5 & 25.2  & 17.1 & 20.3  & 10.9 & 80.0 & 26.4 & 70.2 & 47.1 & 8.4  & 81.5 & 26.0  & 17.2 & 18.9  & 11.7 & \multicolumn{1}{c|}{1.6}  & 36.1 \\
\multicolumn{1}{l|}{FDA\cite{yang2020fda}}         & 86.1 & 35.1  & 80.6  & 30.8 & 20.4  & 27.5 & 30.0  & 26.0 & 82.1 & 30.3 & 73.6 & 52.5 & 21.7 & 81.7 & 24.0  & 30.5 & {\bf29.9}  & 14.6 & \multicolumn{1}{c|}{24.0} & 42.2 \\
\multicolumn{1}{l|}{DAST\cite{yu2021dast}}        & 90.5 & 49.2  & {\bf81.9}  & 34.0 & 27.0  & 26.5 & 26.6  & 21.5 & 83.0 & {\bf37.3} & 76.3 & 52.0 & 23.1 & {\bf83.5} & 29.9  & 42.0 & 12.1  & 19.8 & \multicolumn{1}{c|}{25.8} & 44.3 \\ \hline
\multicolumn{1}{l|}{IDPL(ours)}        &{\bf90.7}      & 50.5      & {\bf81.9}      & {\bf35.8}     & {\bf31.0}      & {\bf30.9}     & {\bf32.1}      &{\bf29.8}      &{\bf83.5}      & 36.4     & {\bf81.1}     & {\bf55.4}     & {\bf28.3}     & 82.2     & {\bf31.2}      & {\bf42.6}     & 19.4      & {\bf22.1}     & \multicolumn{1}{c|}{26.6}     & {\bf46.9}     \\ \hline
ResNet-101              &      &       &       &      &       &      &       &      &      &      &      &      &      &      &       &      &       &      &                           &      \\ \hline
\multicolumn{1}{l|}{Source-Only} & 75.8 & 16.8  & 77.2  & 12.5 & 21.0  & 25.5 & 30.1  & 20.1 & 81.3 & 24.6 & 70.3 & 53.8 & 26.4 & 49.9 & 17.2  & 25.9 & 6.5   & 25.3 & \multicolumn{1}{c|}{36.0} & 36.6 \\
\multicolumn{1}{l|}{AdvEnt\cite{Vu_2019_CVPR}}      & 89.4 & 33.1  & 81.0  & 26.6 & 26.8  & 27.2 & 33.5  & 24.7 & 83.9 & 36.7 & 78.8 & 58.7 & 30.5 & 84.8 & 38.5  & 44.5 & 1.7   & 31.6 & \multicolumn{1}{c|}{32.4} & 45.5 \\
\multicolumn{1}{l|}{CBST\cite{Zou_2018_ECCV}}        & 91.8 & 53.5  & 80.5  & 32.7 & 21.0  & 34.0 & 28.9  & 20.4 & 83.9 & 34.2 & 80.9 & 53.1 & 24.0 & 82.7 & 30.3  & 35.9 & 16.0  & 25.9 & \multicolumn{1}{c|}{42.8} & 45.9 \\
\multicolumn{1}{l|}{UIDA\cite{Pan_2020_CVPR}}        & 90.6 & 37.1  & 82.6  & 30.1 & 19.1  & 29.5 & 32.4  & 20.6 & 85.7 & 40.5 & 79.7 & 58.7 & 31.1 & 86.3 & 31.5  & 48.3 & 0.0   & 30.2 & \multicolumn{1}{c|}{35.8} & 46.3 \\
\multicolumn{1}{l|}{SUDA\cite{zhang2022spectral}}        & 91.1 & 52.3  &{82.9}   & 30.1 & 25.7 & 38.0 & {44.9}  & {38.2} & 83.9 & {39.1} & 79.2 & 58.4 & {26.4} & {84.5} & 37.7  & {45.6} & 10.1   & {23.1} & \multicolumn{1}{c|}{36.0} & 48.8 \\
\multicolumn{1}{l|}{CaCo\cite{huang2022category}}        & 91.9 & 54.3  &{82.7}   & 31.7 & 25.0 & 38.1 & {46.7}  & {39.2} & 82.6 & {39.7} & 76.2 & 63.5 & {23.6} & {85.1} & 38.6  & {47.8} & 10.3   & {23.4} & \multicolumn{1}{c|}{35.1} & 49.2 \\
\multicolumn{1}{l|}{DAST\cite{yu2021dast}}        & 92.2 & 49.0  & 84.3  & 36.5 & 28.9  & 33.9 & 38.8  & 28.4 & 84.9 & 41.6 & 83.2 & 60.0 & 28.7 & 87.2 & {\bf45.0}  & 45.3 & 7.4   & {33.8} & \multicolumn{1}{c|}{32.8} & 49.6 \\
\multicolumn{1}{l|}{RPLL\cite{zheng2021rectifying}}        & 90.4 & 31.2  &{85.1}   & 36.9 & 25.6  & 37.5 & {\bf48.8}  & {\bf48.5} & 85.3 & 34.8 & 81.1 & 64.4 & {\bf36.8} & 86.3 & 34.9  & 52.2 & 1.7   & 29.0 & \multicolumn{1}{c|}{44.6} & 50.3 \\
\multicolumn{1}{l|}{PixMatch\cite{melas2021pixmatch}}        & 91.6 & 51.2  &{84.7}   & 37.3 & 29.1  & 24.6 & {31.3}  & {37.2} & 86.5 & {\bf44.3} & 85.3 & 62.8 & {22.6} & {\bf87.6} & 38.9  & {\bf 52.3} & 0.65   & {\bf37.2} & \multicolumn{1}{c|}{\bf50.0} & 50.3 \\
\hline
\multicolumn{1}{l|}{IDPL(ours)}        &{\bf93.4}      &{\bf55.6}       &{\bf85.3}       &{\bf39.2}      &{\bf40.3}       &{\bf40.1}      & 41.7      &41.2      &{\bf87.0}      &42.3      &{\bf87.8}      &{\bf67.8}      &33.1      &85.1      & 42.2      &52.2      &{\bf22.8}       &33.1      & \multicolumn{1}{c|}{40.6}     &{\bf54.2}      \\ \hline
\end{tabular}}
\label{tab4}
\end{table*}

\begin{table*}[]
\caption{\ IoU (\%) comparison for each class on SYNTHIA $\rightarrow$ Cityscapes adaptation, evaluated on Cityscapes validation set.}
\centering
\scalebox{0.6}{
\begin{tabular}{lcccccccccccccccccc}
\hline
\multicolumn{1}{l|}{Method}      & road  & sidewalk & building & wall* & fence* & pole* & light & sign & veg   & sky   & person  & rider  & car   & bus  & mbike  & \multicolumn{1}{c|}{bike}  & \multicolumn{1}{c|}{mIoU}  & mIoU* \\ \hline
VGG-16                  &       &       &       &       &        &       &       &      &       &       &       &       &       &      &       &                            &                            &       \\ \hline
\multicolumn{1}{l|}{Source Only} & 4.7   & 11.6  & 62.3  & 10.7  & 0.0    & 22.8  & 4.3   & 15.3 & 68.0  & 70.8  & 49.7  & 6.4   & 60.5  & 11.8 & 2.6   & \multicolumn{1}{c|}{4.3}   & \multicolumn{1}{c|}{25.4}  & 28.7  \\
\multicolumn{1}{l|}{CBST\cite{Zou_2018_ECCV}}        & 69.6  & 28.7  & 69.5  & {\bf12.1}  & 0.1    & 25.4  & 11.9  & 13.6 & {\bf82.0}  & 81.9  & 49.1  & 14.5  & 66.0  & 6.6  & 3.7   & \multicolumn{1}{c|}{32.4}  & \multicolumn{1}{c|}{35.4}  & 36.1  \\
\multicolumn{1}{l|}{AdvEnt\cite{Vu_2019_CVPR}}      & 67.9  & 29.4  & 71.9  & 6.3   & 0.3    & 19.9  & 0.6   & 2.6  & 74.9  & 74.9  & 35.4  & 9.6   & 67.8  & 21.4 & 4.1   & \multicolumn{1}{c|}{15.5}  & \multicolumn{1}{c|}{31.4}  & 36.6  \\
\multicolumn{1}{l|}{PyCDA\cite{Lian_2019_ICCV}}       & 80.6  & 26.6  & 74.5  & 2.0   & 0.1    & 18.1  & {\bf13.7}  & 14.2 & 80.8  & 71.0  & 48.0  & 19.0  & 72.3  & 22.5 & 12.1  & \multicolumn{1}{c|}{18.1}  & \multicolumn{1}{c|}{35.9}  & 42.6  \\
\multicolumn{1}{l|}{DAST\cite{yu2021dast}}        & {\bf86.1}  & 35.7  & 79.9  & 5.2   & {0.8}    & 23.1  & 0.0   & 6.9  & 80.9  & 82.5  & 50.6  & 19.8  & {\bf79.7}  & 21.9 & 21.3  & \multicolumn{1}{c|}{38.8}  & \multicolumn{1}{c|}{39.6}  & 46.5  \\
\multicolumn{1}{l|}{FDA\cite{yang2020fda}}         & 84.2  & 35.1  & 78.0  & 6.1   & 0.44   & 27.0  & 8.5   & {\bf22.1} & 77.2  & 79.6  & {\bf55.5}  & 19.9  & 74.8  & 24.9 & 14.3  & \multicolumn{1}{c|}{\bf40.7}  & \multicolumn{1}{c|}{40.5}  & -     \\
\hline
\multicolumn{1}{l|}{IDPL(ours)}        &83.2       &{\bf37.4}       &{\bf80.1}       &11.2       &{\bf0.83}        &{\bf27.9}       &7.5       &19.2      &79.3       &{\bf82.8}       &54.4       &{\bf21.6}       &76.6       &{\bf31.4}      & {\bf22.8}      & \multicolumn{1}{c|}{36.8}      & \multicolumn{1}{c|}{\bf42.1}      &{\bf48.7}       \\ \hline
ResNet-101              &       &       &       &       &        &       &       &      &       &       &       &       &       &      &       &                            &                            &       \\ \hline
\multicolumn{1}{l|}{Source-Only} & 36.30 & 14.64 & 68.78 & 9.17  & 0.20   & 24.39 & 5.59  & 9.05 & 68.96 & 79.38 & 52.45 & 11.34 & 49.77 & 9.53 & 11.03 & \multicolumn{1}{c|}{20.66} & \multicolumn{1}{c|}{29.45} & 33.65 \\
\multicolumn{1}{l|}{AdvEnt\cite{Vu_2019_CVPR}}      & 87.0  & 44.1  & 79.7  & 9.6   & 0.6    & 24.3  & 4.8   & 7.2  & 80.1  & 83.6  & 56.4  & 23.7  & 72.7  & 32.6 & 12.8  & \multicolumn{1}{c|}{33.7}  & \multicolumn{1}{c|}{40.8}  & 47.6  \\
\multicolumn{1}{l|}{UIDA\cite{Pan_2020_CVPR}}        & 84.3  & 37.7  & 79.5  & 5.3   & 0.4    & 24.9  & 9.2   & 8.4  & 80.0  & 84.1  & 57.2  & 23.0  & 78.0  & 38.1 & 20.3  & \multicolumn{1}{c|}{36.5}  & \multicolumn{1}{c|}{41.7}  & 48.9  \\
\multicolumn{1}{l|}{CBST\cite{Zou_2018_ECCV}}        & 68.0  & 29.9  & 76.3  & 10.8  & 1.4    & 33.9  & 22.8  & 29.5 & 77.6  & 78.3  & 60.6  & 28.3  & 81.6  & 23.5 & 18.8  & \multicolumn{1}{c|}{39.8}  & \multicolumn{1}{c|}{42.6}  & 48.9  \\
\multicolumn{1}{l|}{SUDA\cite{zhang2022spectral}}        & 83.4  & 36.0  & 71.3  & 8.7  & 0.1    & 26.0  & 18.2  & 26.7 & 72.4  & {80.2}  & 58.4  & 30.8  & 80.6  & 38.7 & 36.1  & \multicolumn{1}{c|}{46.1}  & \multicolumn{1}{c|}{44.6}  & 52.2  \\
\multicolumn{1}{l|}{DAST\cite{yu2021dast}}        & 87.1  & 44.5  & 82.3  & 10.7  & 0.8    & 29.9  & 13.9  & 13.1 & 81.6  & {\bf86.0}  & 60.3  & 25.1  & 83.1  & 40.1 & 24.4  & \multicolumn{1}{c|}{40.5}  & \multicolumn{1}{c|}{45.2}  & 52.5  \\
\multicolumn{1}{l|}{CaCo\cite{huang2022category}}       & {87.4}  & {48.9}  & {79.6}  & 8.8  & 0.2    & 30.1  & 17.4  & 28.3 & 79.9  & 81.2  & {56.3}  & 24.2  & {78.6}  & 39.2 & 28.1  & \multicolumn{1}{c|}{48.3}  & \multicolumn{1}{c|}{46.0}  & 53.6  \\
\multicolumn{1}{l|}{PixMatch\cite{melas2021pixmatch}}       & {\bf92.5}  & {\bf54.6}  & {79.8}  & 4.78  & 0.08    & 24.1  & 22.8  & 17.8 & 79.4  & 76.5  & {60.8}  & 24.7  & {\bf85.7}  & 33.5 & 26.4  & \multicolumn{1}{c|}{\bf54.4}  & \multicolumn{1}{c|}{46.1}  & 54.5  \\
\multicolumn{1}{l|}{CVRN\cite{huang2021cross}}        & 87.5  & 45.5  & {83.5}  & 12.2  & {0.5}    & 37.4  & 25.1  & 29.6 & {\bf85.9}  & {\bf86.0}  & 61.1  & 25.9  & {80.9}  & 34.7 & 33.8  & \multicolumn{1}{c|}{53.5}  & \multicolumn{1}{c|}{49.0}  & 56.6  \\ 
\hline
\multicolumn{1}{l|}{IDPL(ours)}        & 85.1      & 42.2      & {\bf83.6}      & {\bf22.7}      & {\bf3.8}       & {\bf37.9}      & {\bf30.7}      & {\bf34.6}     & 80.3      & 85.5      &{\bf62.1}       &{\bf35.2}       &{85.4}       &{\bf41.1}      &{\bf38.6}       & \multicolumn{1}{c|}{51.5}      & \multicolumn{1}{c|}{\bf51.3}      &{\bf58.1}       \\ \hline
\end{tabular}}
\label{tab5}
\end{table*}

\section{Conclusion}

In this paper, Intra-subdomain adaptation adversarial learning segmentation method based on Dynamic Pseudo Labels(IDPL) is proposed to improve the segmentation effect of difficult categories. Firstly, each instance is considered by iteration, and the threshold of each class is adjusted adaptively according to the class information in the global classes and local instances, thereby generating high-quality pseudo labels; Secondly, the target domain is dynamically split into two subdomains according to the relative proportion of instances; Finally, the class self-attention mechanism is applied between the two subdomains to enable intra-domain adaptive feature alignment at the class level with the help of high-quality pseudo labels. The overall performance of the proposed method outperforms other mainstream methods by significantly improving the performance of the difficult categories.

\subsubsection*{Acknowledgements.} This work is supported by Tianjin Technical Export Project (Grant No. 21YDTPJC00090).

%
%
%

\end{document}